\documentclass{article} 
\usepackage{iclr2027_conference,times}

\usepackage{amsmath,amsfonts,bm}

\def\eqref#1{equation~\ref{#1}}

\def\1{\bm{1}}

\DeclareMathAlphabet{\mathsfit}{\encodingdefault}{\sfdefault}{m}{sl}
\SetMathAlphabet{\mathsfit}{bold}{\encodingdefault}{\sfdefault}{bx}{n}

\usepackage{amssymb}
\usepackage{hyperref}
\usepackage{url}
\usepackage{graphicx}
\usepackage{booktabs}
\usepackage[table]{xcolor}
\definecolor{tablelightblue}{HTML}{E0F2FC}
\usepackage{tabularx}
\usepackage{multirow}
\usepackage{makecell}
\usepackage{array}
\usepackage{wrapfig}
\usepackage[most]{tcolorbox}
\newcolumntype{Y}{>{\centering\arraybackslash}X}

\title{Weave Forcing: Compositional Memory Routing for Interactive Long Video Generation}

\author{Ziyi Wang\textsuperscript{1}, Junchi Yao\textsuperscript{2}, Heqian Qiu\textsuperscript{1}, Wenbo Shi\textsuperscript{1}, Chengjiu Wang\textsuperscript{1}, Jinyang He\textsuperscript{1} \\
\textbf{Binkai Hong\textsuperscript{1}, Hongliang Li\textsuperscript{1}}\thanks{Corresponding author.} \\
\textsuperscript{1}University of Electronic Science and Technology of China (UESTC) \\
\textsuperscript{2}Mohamed bin Zayed University of Artificial Intelligence (MBZUAI)
}

\iclrfinalcopy 
\begin{document}

\maketitle

\begin{abstract}
Recent advances in autoregressive video generation have improved temporal consistency over extended durations,
yet interactive storytelling requires more than continuous scene extension: a new shot may
combine characters and backgrounds from different historical shots. Whole prompt retrieval can
overlook the distinct reference needs of individual components, while directly combining all historical memories
may introduce unrelated visual content. To address these problems, we present \textbf{Weave Forcing}, a training-free framework for 
compositional memory reuse in interactive long video generation. First, we use an LLM for semantic slot routing to
decompose user prompts into character and background descriptions and explicitly select suitable
historical references for each component. To isolate the required content, masked memory weaving uses contrasting
attention maps conditioned on semantic slots to construct refined semantic masks, selectively exposing relevant tokens from compressed 
historical KV memories to guide the generation of the current shot. We further introduce coverage adaptive RoPE to adjust temporal offsets and memory
retention according to no, partial, or full reference coverage, addressing visual artifacts observed when
incomplete historical references are positioned close to the current generation. 
Extensive experiments demonstrate that Weave Forcing improves cross-shot subject and background consistency 
while maintaining competitive visual quality and text alignment.

\end{abstract}

\addtocontents{toc}{\protect\setcounter{tocdepth}{-1}}
\section{Introduction}
\label{sec:introduction}

Video diffusion models have rapidly advanced visual realism,
motion complexity, and text alignment~\citep{hunyuanvideo,wan}.
Autoregressive models further enable efficient, temporally consistent
long video generation through sequential prediction, historical
context reuse, and improved training and denoising~\citep{nova,selfforcing}.
However, filmmaking and interactive storytelling require users to
revise instructions, switch shots, and recall earlier characters or
backgrounds. Interactive multi-shot generation must therefore
respond to new instructions while preserving continuity with
relevant historical content~\citep{videodirectorgpt}.

Existing approaches can be distinguished by the temporal attention
structure of their video generators. Methods built on bidirectional
diffusion transformers (DiTs) produce high-quality clips and support composition across shots
through joint generation or conditioning on historical frames.
However, repeatedly denoising each target clip with bidirectional
attention incurs substantial computational cost and high generation
latency, limiting rapid user feedback~\citep{longcontexttuning,filmweaver}.
Causal autoregressive approaches instead generate sequentially and
reuse historical KV states; combined with few-step sampling, they
reduce latency while preserving long-range context and motion
dynamics~\citep{causvid,causalcine}.
Within this causal paradigm, global historical context and local
within-shot context, together with shot-transition control, further
support cross-shot consistency and interactive video
generation~\citep{rewardforcing,shotstream}.
These advances improve interactive generation, but precise reuse of
historical content for each new shot remains unresolved.

\begin{figure}[t]
    \centering
    \includegraphics[page=1,width=0.85\linewidth]{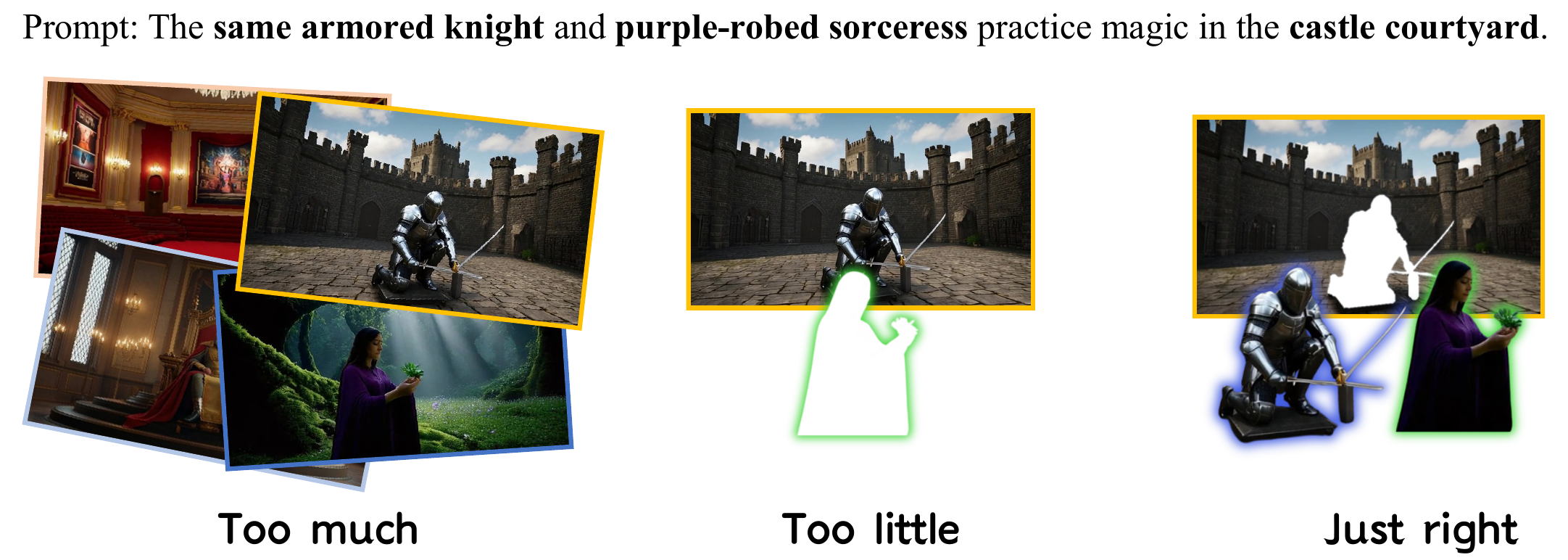}
    \caption{
        \textbf{Motivation for compositional memory reuse.}
        Broad history reuse introduces irrelevant content (too much),
        while global prompt similarity retrieval can miss required references
        (too little). Weave Forcing selectively reuses relevant historical content
        (just right).
    }
    \vspace{-12pt}
    \label{fig:motivation}
\end{figure}

Two limitations arise when instructions combine content from different narrative moments, as shown in Figure~\ref{fig:motivation}.
First, historical references can provide too much
unrelated information or too little relevant evidence. Reusing compressed or sampled
representations of preceding frames preserves whole-scene content,
including unrequested contents~\citep{storymem}.
Conversely, selecting a single source by global prompt similarity,
as in the optional automatic router of Echo Forcing, can miss
required components across
shots~\citep{echoforcing}.
For example, an instruction may bring a previously seen girl and boy
into a library, with each character and the background originating
from a different shot. Reusing all three source memories also imports
unrelated content, whereas the most similar single shot may
omit a required reference. Additional sources alone cannot
resolve within-shot selection.
Second, a uniform reference use policy struggles to reconcile two
competing demands: preserving content supported by historical
references and generating new content without such support.
Strong historical conditioning can constrain unreferenced components,
whereas indiscriminately weakening it can compromise the consistency
of referenced characters and backgrounds. Infinity-RoPE demonstrates
that manipulating temporal rotary position embedding (RoPE) coordinates can regulate temporal
continuity and enable scene transitions~\citep{infinityrope}.
Inspired by this positional control, we adapt temporal RoPE settings
to the availability of character and background references, balancing
reference preservation with flexible generation of new content.

To address these challenges, we propose \textbf{Weave Forcing}, a
training-free framework for compositional memory reuse in interactive
long video generation. Semantic slot routing uses a large language model (LLM) to
first perform slot prompt splitting to identify the requested
characters and background, and then apply cross-shot routing
to associate each component with a historical source and its
corresponding target slot prompt. Masked memory weaving uses
contrastive slot attention (CSA) to distinguish target content from
other content within the same historical memory, producing a
semantic KV mask that selectively exposes relevant tokens.
The references are combined in the cross-shot cache to
guide the new shot. Coverage adaptive RoPE then
adapts temporal positioning and reference retention to the available
character and background support, while progressive memory decay
smooths the transition toward the current shot's own context.
Together, these components enable selective recombination of
historical content as user instructions evolve. Experiments on
100 six-shot prompts demonstrate the effectiveness of
Weave Forcing, showing improved cross-shot subject and background
consistency over the Self Forcing backbone while achieving
competitive visual quality and text alignment.

Our main contributions are summarized as follows:

\begin{itemize}
\item We introduce \textbf{Semantic Slot Routing},
which decomposes each shot prompt into character and background
slots and independently matches them to available historical source
slots, enabling composition of references from different shots.

\item We propose \textbf{Masked Memory Weaving}, which constructs
semantic KV masks through CSA, then combines
the independently masked historical KV memories to retain requested content while suppressing unrelated
characters and backgrounds.

\item We develop \textbf{Coverage Adaptive RoPE}, which
adapts temporal offsets and reference retention to no, partial,
or full character and background reference coverage, balancing
historical guidance with the generation of newly requested content.
\end{itemize}

\section{Related Work}
Early work on autoregressive video generation connected causal sequence
modeling with diffusion through
independent per-token noise levels, enabling generation beyond the
training horizon~\citep{diffusionforcing}. Subsequent approaches
reduced sampling cost through distillation or adversarial
post-training~\citep{causvid,aapt}, while training on self-generated
histories and rolling-window denoising helped reduce error
accumulation~\citep{selfforcing,rollingforcing}. Complementary
system-level optimizations improve the efficiency of streaming
deployment and long video training~\citep{streamdiffusionv2,longlive2}.

Interactive generation additionally requires responsiveness to
changing instructions while maintaining temporal coherence.
Chunk-wise prompting controls successive segments~\citep{magi1}, while
LongLive~\citep{longlive} refreshes cached states under new prompts
to improve transitions. Other work adapts positional encoding and
cache management for extended rollouts and scene
transitions~\citep{infinityrope}, or constrains teacher supervision
to balance responsiveness with stability~\citep{deltaforcing}.
Extending interaction to multi-shot narratives further requires
memory across shot boundaries: global and local contexts support
streaming generation across shots~\citep{shotstream}, and
Echo Forcing~\citep{echoforcing} uses hierarchical memory for scene
recall. In contrast, our work addresses compositional recall by routing character
and background slots to historical sources, masking unrelated KV
content, and adapting memory positioning.

\section{Preliminaries}

\subsection{Autoregressive Long Video Generation}

Autoregressive video diffusion generates a video progressively as
latent blocks, conditioning each block on the text instruction and
previously generated content~\citep{causvid,selfforcing}. Let
$\mathbf{x}=\{\mathbf{x}_b\}_{b=1}^{B}$ denote a sequence of $B$
latent blocks and $P$ its text prompt. The generation distribution
factorizes as
\begin{equation}
    p_{\theta}(\mathbf{x}\mid P)
    =\prod_{b=1}^{B}
    p_{\theta}(\mathbf{x}_b\mid P,\mathbf{x}_{<b}).
\end{equation}
Each conditional distribution is realized by iteratively denoising~\citep{ddpm}
the current block with a causal DiT~\citep{dit}. Historical
key--value (KV) states are cached and reused during attention,
avoiding repeated computation over completed blocks. Once a block
is generated, its clean KV states are added to the cache to support
subsequent generation. This incremental process enables streaming
output and extension beyond a fixed clip duration.

\subsection{3D RoPE}

Video DiTs use 3D RoPE to
encode temporal and spatial token coordinates~\citep{wan,roformer}.
For a token at $\mathbf{p}=(p^{\mathrm{t}},p^{\mathrm{h}},p^{\mathrm{w}})$,
the coordinates specify its temporal, height, and width positions.
Query and key channels are partitioned into three subspaces, each
transformed by axis-specific two-dimensional rotations whose angles
depend on the coordinate and channel frequency. The combined
rotation, transformed queries and keys, and resulting relative position
relation are expressed together as
\begin{equation}
    \begin{gathered}
        \mathbf{R}(\mathbf{p})
        =\operatorname{diag}\!\left(
            \mathbf{R}_{\mathrm{t}}(p^{\mathrm{t}}),
            \mathbf{R}_{\mathrm{h}}(p^{\mathrm{h}}),
            \mathbf{R}_{\mathrm{w}}(p^{\mathrm{w}})
        \right),\\
        \widetilde{\mathbf{q}}_{\mathbf{p}}
        =\mathbf{R}(\mathbf{p})\mathbf{q}_{\mathbf{p}},
        \qquad
        \widetilde{\mathbf{k}}_{\mathbf{p}'}
        =\mathbf{R}(\mathbf{p}')\mathbf{k}_{\mathbf{p}'},\\
        \widetilde{\mathbf{q}}_{\mathbf{p}}^{\top}
        \widetilde{\mathbf{k}}_{\mathbf{p}'}
        =\mathbf{q}_{\mathbf{p}}^{\top}
        \mathbf{R}(\mathbf{p}'-\mathbf{p})\mathbf{k}_{\mathbf{p}'}.
    \end{gathered}
\end{equation}
Thus, the rotary contribution to the attention score depends on the
relative spatiotemporal displacement. Temporal repositioning changes
this relation without modifying spatial coordinates: applying an
offset $J$ to a historical token gives
$\widehat{\mathbf{p}}'=(p^{\mathrm{t}\prime}-J,
 p^{\mathrm{h}\prime},p^{\mathrm{w}\prime})$.
A uniform offset preserves the memory's internal relative geometry
while changing its temporal phase relative to the current queries.
Such coordinate adjustments provide an inference-time means of
controlling temporal relationships, as illustrated by Infinity-RoPE's
use of temporal repositioning and discontinuities for long video generation
and scene transitions~\citep{infinityrope}.

\section{Method}
\label{sec:method}

\subsection{Semantic Slot Routing}
\label{sec:slot-routing}

\begin{figure}[t]
    \centering
    \includegraphics[width=\linewidth]{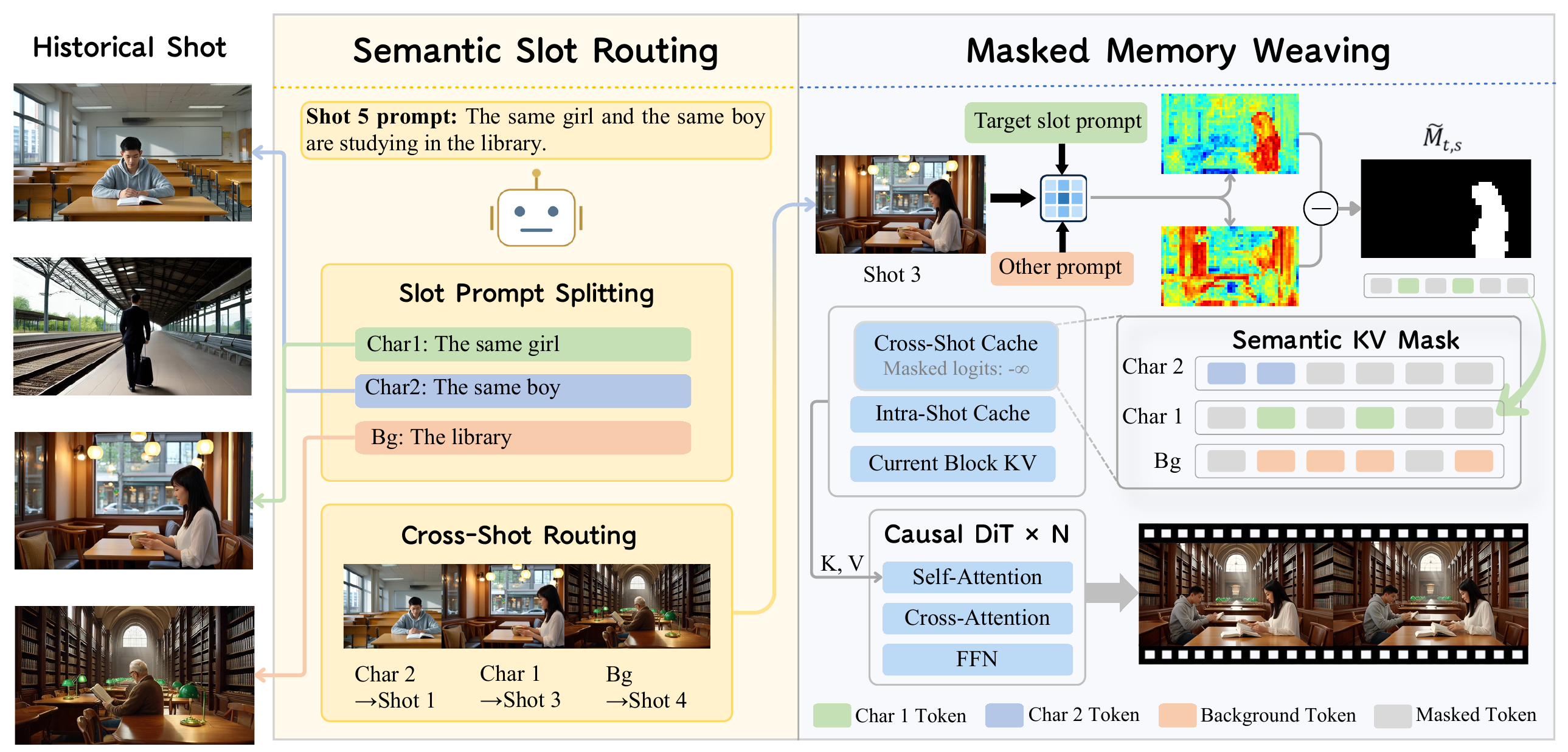}
    \caption{\textbf{Overview of Weave Forcing.}
    Left: Semantic Slot Routing uses slot prompt splitting and cross-shot
    routing to get historical sources. Right: Masked Memory Weaving employs
    contrastive slot attention (CSA) to selectively retain relevant historical tokens and integrates
    the masked memoriesal tokens and integrates the masked memories.}
    \label{fig:slot-routing}
    \vspace{-4pt}
\end{figure}

Cross-shot reference reuse requires identifying the historical content
needed by the current shot. Retrieval based on whole prompt similarity
can miss characters and backgrounds originating from different shots.
Selecting the top-$k$ shots alone does not resolve this issue, since
global similarity neither assigns sources to individual components nor
specifies which content to retain. We therefore use an LLM to establish
these correspondences: slot prompt splitting extracts character
and background descriptions, and cross-shot routing matches each
component to a historical source and its target slot prompt as shown in Figure~\ref{fig:slot-routing} (left). This
assignment supports composition across shots and identifies the relevant
content within each source.

\paragraph{Slot Prompt Splitting.}
Let $P_t$ denote the instruction for shot $t$.
The LLM processes $P_t$ independently and decomposes it
into character and background prompts:
\begin{equation}
    (\{p_{t,\mathrm c_i}\}_{i=1}^{N_t},p_{t,\mathrm b})
    =\operatorname{Split}_{\mathrm{LLM}}(P_t).
\end{equation}
Here, $N_t$ is the number of visible characters, while $\mathrm c_i$ and
$\mathrm b$ denote the $i$-th character and background, respectively.
Each $p_{t,\mathrm c_i}$ retains the identity and appearance attributes
of one character, while $p_{t,\mathrm b}$ describes the environment. This step yields
separate character and background descriptions that serve as the
semantic queries for subsequent source routing.

\paragraph{Cross-Shot Routing.}
Let $\mathcal H_{t-1}$ denote the descriptions and indices of preceding
shots. The LLM independently matches each
character or background slot to an eligible historical source:
\begin{equation}
    j_{t,s}=\operatorname{Route}_{\mathrm{LLM}}
    (p_{t,s},\mathcal H_{t-1}),
    \qquad s\in\{\mathrm c_1,\ldots,\mathrm c_{N_t},\mathrm b\}.
\end{equation}
Here, $j_{t,s}$ identifies the source shot for slot $s$. Only shots
preceding $t$ are eligible as sources. The LLM identifies the matching slot within that source, whose extracted
description serves as the target slot prompt. The source index locates the
compressed KV memory, while the target slot prompt specifies the
character or background content to retain. If no suitable source
exists, $j_{t,s}=\varnothing$ and
no reference is assigned. All detailed prompt instructions are provided
in Appendix~\ref{app:prompts}.

\subsection{Masked Memory Weaving}
\label{sec:masked-weaving}

Historical shots selected by the router can contain relevant
characters or backgrounds alongside unrelated content. Directly
reusing their complete memories may therefore introduce visual
information that conflicts with the current instruction. We address
this interference through slot-specific semantic KV masks that isolate
the requested content within each retrieved memory. At the beginning
of a new shot, the masked historical tokens form KV references that
guide the first generated blocks, as illustrated in
Figure~\ref{fig:slot-routing} (right).

\paragraph{Memory compression with query weighting.}
To make references available at shot boundaries, we compress the KV representations of each completed shot $j$ into a compact memory. 
Let $\bar Q_j$ be the mean query over the shot. 
For each spatial token $u$, we normalize its query--key affinities across
candidate frames $f$ and aggregate the K/V pairs using the
weights:
\begin{equation}
    \alpha_{j,f,u}=\operatorname{softmax}_{f}
    \!\left(\langle \bar Q_j,K_{j,f,u}\rangle\right),\qquad
    (\bar K_{j,u},\bar V_{j,u})=
    \sum_f\alpha_{j,f,u}(K_{j,f,u},V_{j,f,u}).
\end{equation}
The memory is indexed by shot for retrieval via $j_{t,s}$.
\paragraph{Contrastive slot attention (CSA).}
A target prompt alone can activate both the requested component and
its correlated surroundings. We use CSA to discriminate the target from competing content
through attention map differencing.
For each character or background slot $s$, let $j_{t,s}$ denote its selected source shot. For slots with a valid source,
we use two complementary textual conditions:
the target prompt $P_{t,s}^{\mathrm{tar}}$ specifies the
content to retain, while the other prompt
$P_{t,s}^{\mathrm{oth}}$ describes non-target content to suppress
within the selected source (Appendix~\ref{app:routing-shot-prompts}). Both prompts are evaluated on the same historical
memory, yielding aligned token-level attention maps:
\begin{align}
    A_{t,s}^{\mathrm{tar}}
    &=\mathcal{A}_{\mathrm{cross}}
      \left(P_{t,s}^{\mathrm{tar}},\bar K_{j_{t,s}}\right),\\
    A_{t,s}^{\mathrm{oth}}
    &=\mathcal{A}_{\mathrm{cross}}
      \left(P_{t,s}^{\mathrm{oth}},\bar K_{j_{t,s}}\right).
\end{align}
We subtract the other content response from the target response and
threshold the resulting contrastive attention score to obtain
an initial binary mask $M_{t,s}$:
\begin{equation}
    M_{t,s,u}=\mathbb{I}\!\left[
        \left(A_{t,s,u}^{\mathrm{tar}}-A_{t,s,u}^{\mathrm{oth}}\right)>\tau_s
    \right].
    \label{eq:contrastive-slot-attention}
\end{equation}
Here, $u$ indexes a token in the shared memory layout and $\tau_s$
is the threshold for slot $s$. The contrast is computed at inference time and
introduces no contrastive training objective. For example, differencing the heatmaps
for the girl and the other content highlights the girl while
suppressing the library surroundings.
The initial binary mask is reshaped to the spatial grid associated with the compressed memory and refined using morphological opening and closing~\citep{soille1999} to obtain the final semantic KV mask $\widetilde M_{t,s}$, as follows:
\begin{equation}
    \widetilde M_{t,s}
    =\mathcal{F}_{\mathrm{morph}}(M_{t,s})
    =\operatorname{Close}\!\left(\operatorname{Open}(M_{t,s})\right).
\end{equation}
Opening removes isolated high-response regions, whereas closing fills small holes and connects nearby target regions. In combination, these operations reduce spurious activations and improve the spatial coherence of the mask, supporting more complete retention of the target content. The semantic KV mask $\widetilde M_{t,s}$ is then mapped back to the memory token layout to control which historical KV tokens are accessible to the current block.

\paragraph{Cache initialization with semantic masks.}
At shot $t$, we reset the context and load compressed memories from
valid source indices, retaining independently masked views for slots
sharing a source. The bounded context has four windows:
sink initially stores historical references, then rolling anchors;
compressed retains compressed within-shot history;
recent holds the last completed block; and current
holds the block being denoised. This layout treats cross-shot references
as temporary contents, not a separate window. The current block supplies
queries, while occupied window entries supply keys and values:
\begin{equation}
    \begin{aligned}
    K_{\mathrm{all}}&=[K_{\mathrm{sink}};K_{\mathrm{comp}};K_{\mathrm{recent}};K_{\mathrm{cur}}],\\
    V_{\mathrm{all}}&=[V_{\mathrm{sink}};V_{\mathrm{comp}};V_{\mathrm{recent}};V_{\mathrm{cur}}].
    \end{aligned}
\end{equation}
In the Causal DiT, self-attention~\citep{attention} integrates these states
and gates each loaded cross-shot view with its slot-specific mask
$\widetilde M_{t,s}$. The semantic bias is zero for all current-shot
tokens, including subsequent sink entries, rather than permanently
masking the sink window. For loaded reference tokens:
\begin{equation}
    \begin{gathered}
        O_t=\operatorname{Softmax}\!\left(
            \frac{Q_tK_{\mathrm{all}}^{\top}}{\sqrt d}
            +B_{\mathrm{sem}}
        \right)V_{\mathrm{all}},\\
        B_{\mathrm{sem}}(q,s,u)=
        \begin{cases}
            0, & \widetilde M_{t,s,u}=1,\\
            -\infty, & \widetilde M_{t,s,u}=0.
        \end{cases}
    \end{gathered}
\end{equation}
This masked attention supplies relevant cross-shot references to
guide generation as the current shot develops its own context.

\noindent
\begin{minipage}{\textwidth}
\expandafter\def\csname @captype\endcsname{figure}
    \centering
    \includegraphics[page=3,width=\linewidth]{fig/fig.pdf}
    \caption{\textbf{Coverage adaptive RoPE and cache evolution.}
    Top: no coverage. Middle: partial coverage. Bottom: full coverage.}
    \label{fig:coverage-rope}
\end{minipage}
\par

\subsection{Coverage Adaptive RoPE}
\label{sec:coverage-rope}

As the current shot accumulates context, historical guidance must be
regulated. Following the transient reference reuse of
Echo Forcing~\citep{echoforcing}, we remove loaded cross-shot references
after a prescribed number of completed blocks, rather than upon reaching
the window capacity: one block for no coverage and five for partial or
full coverage in our configuration. This eviction frees space for
within-shot history without removing the sink window itself.
The sink, compressed, and recent windows maintain rolling anchors,
compressed history, and recent context, respectively, while the current
window holds the block being denoised. Their updates support the
transition from historical guidance to the current shot's own context.
The similarity-adaptive decay rule is detailed in
Appendix~\ref{app:similarity-adaptive-decay}.

However, a uniform retention policy is insufficient because referenced
and unreferenced content require different positional treatment. We therefore
distinguish no coverage, partial coverage, and
full coverage according to character and background reference
availability, with corresponding temporal RoPE configurations and
retention policies (Figure~\ref{fig:coverage-rope}).

\noindent\textbf{No coverage.}
When both character and background reference sets are empty, we
use the preceding shot's complete compressed memory as a transient
style reference without semantic masking. We assign a temporal RoPE
offset of $45$ to separate it from the current queries and remove it
after the first block. Subsequent generation relies on the Intra-Shot
Cache, preserving stylistic continuity while limiting unrelated
historical influence.

\noindent\textbf{Partial coverage.}
When only one semantic category has references, we insert the
corresponding masked memories into the Cross-Shot Cache.
Since uneven reference support across semantic components may cause
drift, we use a temporal RoPE offset of $45$ to increase its
positional separation from the current queries. The reference then
decays progressively until eviction, preserving guidance for the
supported component while allowing missing content to develop from
the current prompt.

\noindent\textbf{Full coverage.}
When both semantic categories have references, we use their masked
memories with similarity-dependent temporal RoPE offsets.
We set the offset within $[1,6]$ based on the similarity between
the reference and current-shot prompt embeddings. This close positional
coupling facilitates the joint reuse of character and background
references, while progressive decay before eviction supports the
transition to the current shot's own context.

\section{Experiments}

\paragraph{\textbf{Dataset.}}
Using Gemini 2.5 Pro, we construct 100 test narratives by generating
concise prompts and expanding each into six detailed shot prompts.
Each shot lasts 10 seconds, yielding 60-second videos.
Appendix~\ref{app:dataset-prompts} provides the expansion template.

\paragraph{\textbf{Implementation Details.}}
We build Weave Forcing upon Self Forcing~\citep{selfforcing} with the
Wan2.1-T2V-1.3B backbone~\citep{wan} and use Gemini 2.5 Pro~\citep{gemini25pro}
for slot prompt decomposition and causal cross-shot reference
routing. We generate videos at a resolution of $832\times480$ pixels
with random seed $0$ and configured denoising timesteps
$[1000,750,500,250]$.
The model generates three latent frames per autoregressive block,
with a total KV cache capacity of 21 latent frames, allocated to
the sink, compressed, recent, and current windows with capacities
of 12, 3, 3, and 3 latent frames, respectively.
All experiments are conducted on an NVIDIA RTX PRO 6000 GPU
with 96\,GB of memory.

\paragraph{\textbf{Evaluation Metrics.}}
We evaluate subject and background consistency at both
intra-shot and inter-shot levels.
We sample frames separately for intra-shot and inter-shot
evaluation. We then use YOLO~\citep{redmon2016yolo} for subject localization
and SAM~\citep{kirillov2023sam} for segmentation, separating subject regions.
DINOv2~\citep{oquab2023dinov2} features are then extracted from the corresponding
regions and compared using cosine similarity to measure visual
consistency.
We use VBench-Long~\citep{huang2024vbench,huang2025vbenchpp} to assess
aesthetic and imaging quality on each complete video. For text alignment,
we apply the VBench ViCLIP metric~\citep{wang2023internvid} to each shot and
its corresponding short raw prompt (Appendix~\ref{app:metrics}).
These metrics jointly characterize visual quality, prompt
alignment, and consistency across temporal scales.
\begin{table}[htbp]
\centering
\small
\caption{\textbf{Quantitative comparison with existing methods.}
$\uparrow$ indicates that higher is better.
Sub., Bg., and Cons. denote subject, background, and consistency,
respectively. Best results are in \textbf{bold}, and second-best results
are \underline{underlined}.}
\label{tab:quantitative-comparison}

\setlength{\tabcolsep}{3pt}
\renewcommand{\arraystretch}{1.1}

\begin{tabularx}{\textwidth}{
    @{}l*{4}{>{\hsize=0.775\hsize\linewidth=\hsize\centering\arraybackslash}X}
    >{\hsize=1.25\hsize\linewidth=\hsize\centering\arraybackslash}X
    >{\hsize=1.2\hsize\linewidth=\hsize\centering\arraybackslash}X
    >{\hsize=1.45\hsize\linewidth=\hsize\centering\arraybackslash}X@{}
}
\toprule
\multirow{2}{*}{Method}
& \multicolumn{2}{c}{Intra-shot Cons.}
& \multicolumn{2}{c}{Inter-shot Cons.}
& \multirow{2}{*}{\makecell{Aesthetic\\Quality$\uparrow$}}
& \multirow{2}{*}{\makecell{Imaging\\Quality$\uparrow$}}
& \multirow{2}{*}{\makecell{Text\\Alignment$\uparrow$}} \\
\cmidrule(lr){2-3}\cmidrule(lr){4-5}
& Sub.$\uparrow$
& Bg.$\uparrow$
& Sub.$\uparrow$
& Bg.$\uparrow$
& & & \\
\midrule

\rowcolor{tablelightblue}
\multicolumn{8}{l}{\textit{Bidirectional models}} \\
EchoShot~\citep{echoshot} & 0.776 & \underline{0.950} & 0.338 & 0.652 & 0.594 & 0.687 & 0.231 \\
CineTrans~\citep{cinetrans} & \textbf{0.845} & \textbf{0.983} & 0.354 & 0.652 & \underline{0.609} & 0.712 & 0.227 \\
\midrule
\rowcolor{tablelightblue}
\multicolumn{8}{l}{\textit{Autoregressive models}} \\
Self Forcing~\citep{selfforcing} & 0.702 & 0.878 & 0.405 & 0.648 & 0.590 & 0.682 & 0.160 \\
LongLive~\citep{longlive} & 0.758 & 0.894 & 0.489 & \underline{0.769} & 0.596 & 0.734 & 0.182 \\
Rolling Forcing~\citep{rollingforcing} & \underline{0.814} & 0.941 & 0.435 & 0.671 & 0.598 & \underline{0.748} & \textbf{0.239} \\
Infinity-RoPE~\citep{infinityrope} & 0.805 & 0.903 & \underline{0.495} & 0.735 & 0.589 & 0.732 & 0.209 \\
Reward Forcing~\citep{rewardforcing} & 0.719 & 0.845 & 0.472 & 0.749 & 0.590 & \textbf{0.750} & 0.176 \\
ShotStream~\citep{shotstream} & 0.802 & 0.885 & 0.478 & 0.662 & 0.595 & 0.705 & 0.218 \\
Echo Forcing~\citep{echoforcing} & 0.799 & 0.935 & 0.431 & 0.708 & 0.605 & 0.731 & 0.227 \\
\midrule
\textbf{Ours} & 0.771 & 0.923 & \textbf{0.513} & \textbf{0.811} & \textbf{0.611} & \textbf{0.750} & \underline{0.237} \\

\bottomrule
\end{tabularx}
\end{table}
\subsection{Quantitative Results}

Table~\ref{tab:quantitative-comparison} compares Weave Forcing with
representative bidirectional and autoregressive baselines. Bidirectional
models exhibit stronger intra-shot consistency, likely benefiting from
access to the full shot context when modeling dependencies among frames
during generation. Although Weave Forcing does not lead in intra-shot
consistency, it achieves the highest inter-shot subject and background
consistency among the compared methods, demonstrating its strength in
preserving both characters and backgrounds across shot boundaries.
This supports the effectiveness of selectively reusing cache components
relevant to the current instruction, rather than relying on a single
historical source or uniformly sampled historical frames. Alongside
these cross-shot consistency gains, it maintains competitive video
quality, including imaging quality and text alignment. We attribute
this in part to our coverage adaptive RoPE strategy, which adjusts the use and
progressive decay of historical references according to the available
character and background support, helping mitigate visual drift and
unstable transitions. Overall, the quantitative results demonstrate
the effectiveness of selective cross-shot memory reuse in preserving
historical content while maintaining competitive visual quality and
alignment with the current prompt.
\begin{figure}[h]
    \centering
    \includegraphics[page=4,width=\linewidth]{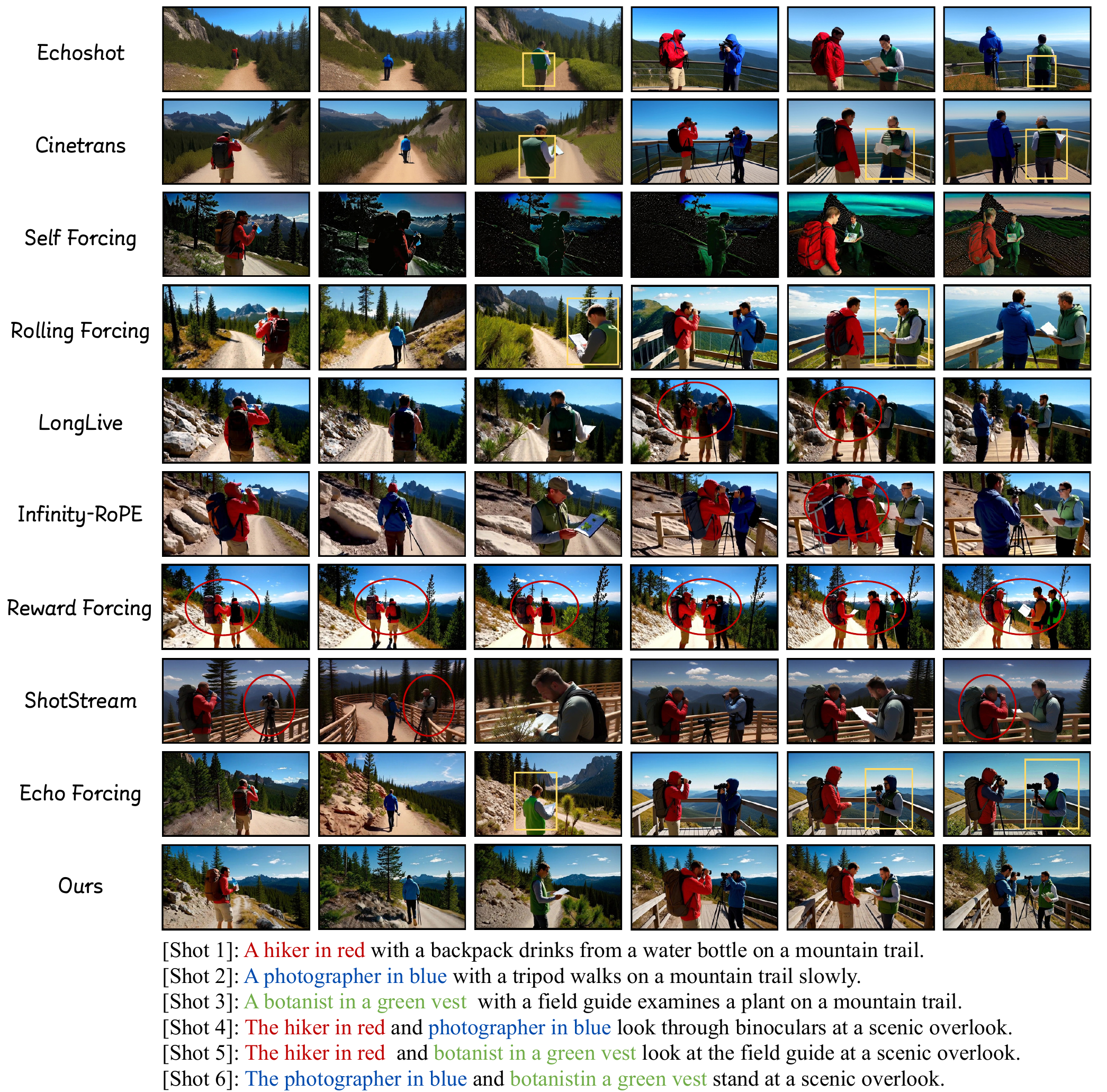}
    \caption{\textbf{Qualitative comparison across six sequential shots.}
    Red circles (\textcolor{red}{$\bigcirc$}) mark prompt mismatches, and
    yellow boxes (\textcolor{yellow}{$\square$}) mark inconsistencies
    with earlier shots.}
    \label{fig:qualitative-comparison}
    \vspace{-4pt} 
\end{figure}
\subsection{Qualitative Results}

Figure~\ref{fig:qualitative-comparison} shows a representative
interactive generation example. The first three shots establish individual
characters, while later prompts require switching scenes and combining
different characters. Self Forcing does not explicitly support free
shot-level switching; in this example, its generated frames are dark,
making their visual content difficult to discern. Methods such as
LongLive can generate content that clearly deviates from the prompt,
as marked by the red circles. We further observe that methods such as
EchoShot and Rolling Forcing may match the
requested scene but still alter referenced details such as clothing
materials or trouser colors, as highlighted by the yellow boxes. This
suggests that they may rely more on prompt-level matching than on faithful
reuse of earlier visual content.
In contrast, Weave Forcing better
preserves these appearance details when composing characters drawn from different
historical shots. This behavior is consistent with semantic slot routing and
masked memory reuse, which provide relevant visual references while
limiting interference from unrelated content.
Additional qualitative results and analyses are provided in
Appendix~\ref{app:additional-results} and Appendix~\ref{app:limitations}.

\subsection{Ablation Study}

To verify the contribution of each component, we conduct ablation studies
over cache fusion, reference routing, semantic isolation, cross-shot
guidance duration, and intra-shot history retention.

\begin{figure}[h]
\noindent
\begin{minipage}{\textwidth}
\centering
\makeatletter
\def\@captype{table}
\makeatother
\newsavebox{\fusiontablebox}
\sbox{\fusiontablebox}{%
\footnotesize%
\setlength{\tabcolsep}{0.5pt}%
\renewcommand{\arraystretch}{1.15}%
\begin{tabularx}{0.51\textwidth}{@{}l*{5}{>{\centering\arraybackslash}X}@{}}
\toprule
\multirow{2}{*}{Variant}
& \multicolumn{2}{c}{Intra-shot$\uparrow$}
& \multicolumn{2}{c}{Inter-shot$\uparrow$}
& \multirow{2}{*}{\makecell{Text\\Align.$\uparrow$}} \\
\cmidrule(lr){2-3}\cmidrule(lr){4-5}
& Sub. & Bg. & Sub. & Bg. & \\
\midrule
Concat & 0.778 & 0.929 & 0.493 & 0.809 & 0.224 \\
Weighted & \textbf{0.793} & \textbf{0.933} & 0.414 & 0.687 & 0.228 \\
\midrule
\textbf{Ours} & 0.771 & 0.923 & \textbf{0.513} & \textbf{0.811} & \textbf{0.237} \\
\bottomrule
\end{tabularx}%
}
\newsavebox{\fusionpanelbox}
\newsavebox{\fusioncaptionbox}
\sbox{\fusionpanelbox}{%
\begin{minipage}{0.51\textwidth}
\centering
\setlength{\abovecaptionskip}{0pt}
\caption{\textbf{Cross-shot cache fusion strategy.} Align. denotes alignment. Best results are in \textbf{bold}.}
\label{tab:ablation}
\usebox{\fusiontablebox}
\end{minipage}%
}
\sbox{\fusioncaptionbox}{%
\begin{minipage}{0.45\textwidth}
\centering
\expandafter\def\csname @captype\endcsname{figure}
\setlength{\abovecaptionskip}{6pt}
\caption{\textbf{Qualitative comparison} of weighted fusion, concatenation, and Ours.}
\label{fig:ablation-qualitative}
\end{minipage}%
}
\noindent
\raisebox{-\height}{\usebox{\fusionpanelbox}}\hfill
\raisebox{-\height}{\vbox{%
\hbox{\includegraphics[page=5,width=0.45\textwidth]{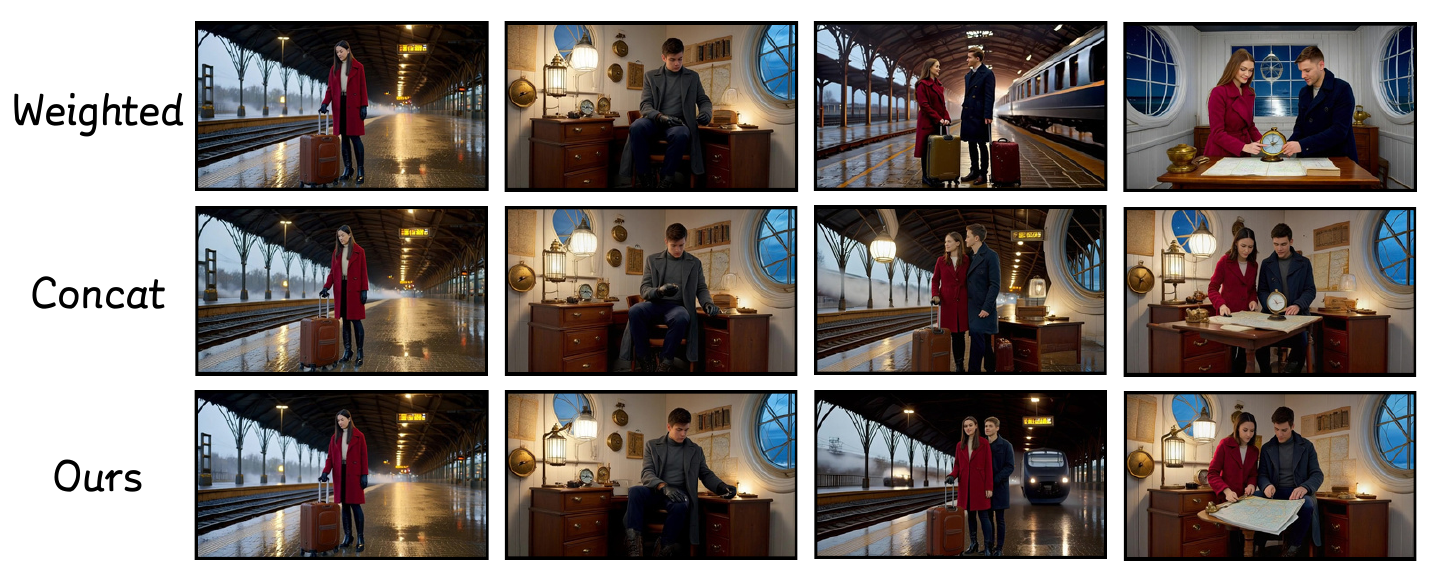}}%
\nointerlineskip
\hbox{\usebox{\fusioncaptionbox}}%
}}
\end{minipage}
\par
\end{figure}

\paragraph{Cross-shot cache fusion strategy.}
Table~\ref{tab:ablation} and Figure~\ref{fig:ablation-qualitative} compare
cache fusion strategies quantitatively and qualitatively. Ours (semantic KV mask) outperforms direct concatenation and weighted fusion in cross-shot
consistency and text alignment. The qualitative example shows that semantic masks reduce 
background mixing and better preserve reference details across shots when 
combining memories from different historical sources, supporting the effectiveness
of our semantic masks.

\paragraph{Individual components.}\par
\begin{wraptable}{l}{0.55\columnwidth}
\vspace{-\intextsep} 
\centering
\setlength{\abovecaptionskip}{0pt}
\caption{\textbf{Component ablations.} Align. denotes alignment.
Best results are in \textbf{bold}.}
\label{tab:ablation-other}
\footnotesize
\setlength{\tabcolsep}{0pt}
\renewcommand{\arraystretch}{1.15}
\resizebox{\linewidth}{!}{%
\begin{tabular}{
    @{}>{\raggedright\arraybackslash}p{0.3\linewidth}
    *{5}{>{\centering\arraybackslash}p{0.14\linewidth}}@{}
}
\toprule
\multirow{2}{*}{Variant}
& \multicolumn{2}{c}{Intra-shot$\uparrow$}
& \multicolumn{2}{c}{Inter-shot$\uparrow$}
& \multirow{2}{*}{\makecell{Text\\Align.$\uparrow$}} \\
\cmidrule(lr){2-3}\cmidrule(lr){4-5}
& Sub. & Bg. & Sub. & Bg. & \\
\midrule
\rowcolor{tablelightblue}
\multicolumn{6}{l}{\textit{Reference routing}} \\
w/o LLM & \textbf{0.790} & \textbf{0.938} & 0.442 & 0.734 & 0.230 \\
\addlinespace[2pt]
\rowcolor{tablelightblue}
\multicolumn{6}{l}{\textit{Semantic separation}} \\
w/o separation & 0.777 & 0.927 & 0.501 & 0.778 & 0.220 \\
\addlinespace[2pt]
\rowcolor{tablelightblue}
\multicolumn{6}{l}{\textit{Memory decay}} \\
w/o decay & 0.765 & 0.919 & 0.471 & 0.780 & 0.219 \\
\addlinespace[2pt]
\rowcolor{tablelightblue}
\multicolumn{6}{l}{\textit{Intra-shot compression}} \\
w/o compression & 0.765 & 0.916 & 0.475 & 0.766 & 0.224 \\
\midrule
\textbf{Ours} & 0.771 & 0.923 & \textbf{0.513} & \textbf{0.811} & \textbf{0.237} \\
\bottomrule
\end{tabular}%
}
\end{wraptable}
Table~\ref{tab:ablation-other} ablates reference routing, semantic isolation,
cross-shot guidance duration, and intra-shot history retention
(implementation details in Appendix~\ref{app:method-details}). Each
component supports stable long-form generation. Without LLM-assisted
routing, cross-shot preservation declines because global prompt similarity
can miss required characters or backgrounds. Removing semantic masking
reduces inter-shot consistency and text alignment, supporting separate
isolation even for components from the same source frame. Without
progressive decay, historical references persist excessively, hindering
the transition to current-shot context. Removing intra-shot compression
discards compact history, weakening both local consistency and cross-shot
preservation. The full method achieves the best inter-shot consistency
and text alignment, highlighting the need to jointly assess intra-shot
and cross-shot criteria.

\section{Conclusion}
We presented Weave Forcing, a training-free framework for
compositional memory reuse in interactive long video
generation.
Semantic slot routing uses an LLM to select historical sources for individual
characters and backgrounds, while CSA constructs
semantic KV masks to combine relevant memory content with reduced
interference from unrelated components. Coverage adaptive RoPE
adjusts temporal offsets and reference retention to the available support,
balancing historical continuity with new content generation as memory
decay shifts guidance toward the current shot's own context.
Experiments demonstrate improved inter-shot subject consistency while maintaining
competitive visual quality and text alignment.
These findings suggest that effective memory reuse for interactive video
generation should prioritize the relevance of individual semantic components
to the current instruction, rather than treating historical scenes as
indivisible visual references.

\subsection*{AI use statement}

In this work, we used generative AI tools for writing assistance and research execution. Specifically, ChatGPT was used to improve the clarity, grammar, and overall presentation of the manuscript. Gemini 2.5 Pro was used to assist the development and implementation of the semantic slot routing component, as well as to generate prompts for constructing the synthetic test datasets used in our experiments.
All AI-assisted outputs were carefully reviewed, verified, and revised by the authors.
We take responsibility for the final content of this work.

\subsection*{Reproducibility statement}

To facilitate reproducibility, Section~\ref{sec:method} describes the proposed
method, while Appendices~\ref{app:prompts}, \ref{app:method-details},
and~\ref{app:metrics} provide the prompt templates, implementation details,
and evaluation procedures, respectively. These details provide
the information needed to reproduce our work.

\bibliography{main}
\bibliographystyle{iclr2027_conference}

\clearpage
\appendix

\addtocontents{toc}{\protect\setcounter{tocdepth}{2}}
\begingroup
\definecolor{appendixnavy}{HTML}{244663}
\definecolor{appendixslate}{HTML}{526C80}
\hypersetup{colorlinks=true,linkcolor=appendixnavy,linktoc=all}
\renewcommand{\contentsname}{\textcolor{appendixnavy}{\fontsize{24pt}{29pt}\selectfont\bfseries Appendix Contents}}
\makeatletter
\let\appendix@section\l@section
\let\appendix@subsection\l@subsection
\renewcommand*{\l@section}[2]{%
  \addvspace{3pt}%
  \begingroup
  \fontsize{12pt}{15pt}\selectfont
  \color{appendixnavy}\hypersetup{linkcolor=appendixnavy}%
  \appendix@section{#1}{#2}%
  \endgroup}
\renewcommand*{\l@subsection}[2]{%
  \addvspace{2pt}%
  \begingroup
  \color{appendixslate}\hypersetup{linkcolor=appendixslate}%
  \appendix@subsection{#1}{#2}%
  \endgroup}
\renewcommand{\tableofcontents}{%
  \section*{\contentsname}%
  \vspace{-20pt}%
  \@starttoc{toc}}
\makeatother
\begin{tcolorbox}[
  enhanced, breakable,
  colback=white, colframe=appendixnavy!65,
  boxrule=0.6pt, arc=2pt,
  left=12pt, right=12pt, top=4pt, bottom=10pt
]
\renewcommand{\baselinestretch}{1.12}\selectfont
\tableofcontents
\end{tcolorbox}
\endgroup
\setcounter{tocdepth}{2}
\clearpage

\section{LLM Prompts and Computational Cost}
\label{app:prompts}

All language-model operations described in this section---multi-shot prompt
expansion, semantic-slot extraction, and cross-shot source routing---are
performed with Gemini 2.5 Pro~\citep{gemini25pro}. The following templates
specify the corresponding inputs, constraints, and structured outputs.

\subsection{Multi-Shot Prompt Expansion}
\label{app:dataset-prompts}

\noindent\textbf{Summary.}
The shot planner converts one raw narrative into chronological, directly
usable video-generation prompts.  It computes $\lceil T/10\rceil$ shots for
a video of length $T$ seconds (six 10-second shots in our 60-second setting),
preserves the order of events, and requires complete repeated descriptions of
recurring characters and locations. The implementation uses the following
template with runtime fields shown in braces.

\begin{tcblisting}{listing only, breakable, colback=gray!4, colframe=gray!60,
  boxrule=0.4pt, arc=1pt, left=4pt, right=4pt, top=4pt, bottom=4pt,
  listing options={basicstyle=\ttfamily\fontsize{8pt}{10pt}\selectfont, breaklines=true}}
You are a cinematic multi-shot video prompt planner. Start from the raw prompt
and expand it into a short sequence of complete, directly usable
video-generation prompts. Stay close to the raw prompt: preserve its main
characters, setting, visual style, and central action. Add only the minimal
details needed for visual continuity between shots. Keep the story simple and
easy to depict. Do not introduce complicated plots, multiple sub-events,
unnecessary characters, dramatic twists, or long chains of actions. Do not
change the identity or defining appearance of an established character. Keep
every shot visually concrete, temporally consistent, and clear, natural, and
grammatically correct English. Return JSON only.

# TASK
Transform the raw prompt into a chronological sequence of complete
video-generation scene prompts.

# RAW PROMPT
{raw_prompt}

# SHOT PLAN
- Split the total duration of {duration_seconds} seconds into {shot_count}
  chronological scene shots.
- Use these exact shot durations: {shot_durations}.
- Use scene1, scene2, ... in chronological order.
- Plan shots strictly in the order expressed in the RAW PROMPT; do not reorder,
  reverse, or otherwise change events, actions, or descriptions.
- Each shot depicts one simple scene: first describe the background, then the
  visible character(s) and their one simple action or stable state.
- Do not use complex action chains, rapid camera changes, or separate actions,
  poses, gestures, reactions, gaze directions, or motion states for different
  visible characters. Multiple characters share the same action or state.

# FOR BACKGROUNDS
In every visible setting, give a full concrete description of the location and
its architecture or natural environment. Repeat it independently in every
shot. On recurrence, begin with 'back in' and repeat the full supported
description; never write only 'the same place' or 'the same'.

# FOR CHARACTERS
On first appearance, give all supported stable details, including approximate
age, gender presentation, clothing colours, and visual traits. On recurrence,
begin with 'the same' and repeat the full supported appearance; never write
only 'the same person', 'the same woman', 'the same man', or 'the same'. Do
not invent unsupported details or change clothing, accessories, hairstyle, or
other defining details. Do not add a second person when the raw prompt has no
two-person shot. A character is fully present or absent: when absent, do not
mention its body part, clothing, shadow, reflection, or off-screen implication.
Use one mostly static state per shot. In a multi-character shot, describe each
individual's appearance in one uninterrupted phrase without commas and do not
interleave attributes across characters.

# GENERAL RULES
- Do not remove supported visual details merely to shorten a shot.
- Keep every shot clear, natural, and directly usable by a video generator.
- Do not add explanations, commentary, or markdown fences.

# OUTPUT FORMAT
Return exactly this JSON shape, with no additional keys or text:
{"shots":[{"shot_id":"scene1","duration_seconds":10,
"prompt":"complete shot prompt"}]}
Each object in "shots" contains only shot_id, duration_seconds, and prompt.
\end{tcblisting}

\subsection{Semantic Slot Extraction}
\label{app:slot-prompts}

\noindent\textbf{Summary.}
For every expanded scene independently, the semantic planner separates the
description into a variable number of character slots and one background
slot. Distinct visible subjects are represented by \texttt{character1},
\texttt{character2}, and so forth; a scene containing only one subject does
not introduce an empty second-character field. This conservative decomposition
retains static identity and appearance cues while excluding actions and poses.
No cross-scene coreference resolution is performed during slot extraction.

\begin{tcblisting}{listing only, breakable, colback=gray!4, colframe=gray!60,
  boxrule=0.4pt, arc=1pt, left=4pt, right=4pt, top=4pt, bottom=4pt,
  listing options={basicstyle=\ttfamily\fontsize{8pt}{10pt}\selectfont, breaklines=true}}
You are a video prompt semantic slot extractor. For each scene prompt, produce
one slot for each distinct visible character, named character1, character2,
and so forth, together with one background slot. Return JSON only and process
each scene independently. Use only information explicitly present in that
scene prompt and retain its original wording whenever possible.

# TASK
Separate each scene prompt into character1, character2, ... and background.

# SCENES
scene1: {scene1_prompt}
scene2: {scene2_prompt}
...

# SLOT EXTRACTION RULES
- Process each scene independently; do not use other scenes.
- Use only information explicit in the current prompt; do not infer, resolve
  references, or invent details.
- Create exactly one characterN slot for each distinct visible person or
  creature. Number slots by order of first mention in the scene.
- Do not output characterN when the corresponding character is absent.
- Each characterN slot contains only that subject's identity, appearance,
  clothing, accessories, and static visual traits; exclude expression, pose,
  movement, behavior, actions, and attributes belonging to another subject.
- background: only place, environment, props, lighting, atmosphere, and camera
  setting as compact noun phrases; exclude people, verbs, and prepositions.
- Keep original wording and visual details whenever possible.

# OUTPUT FORMAT
Return exactly this JSON shape and no additional text:
{"scenes":[{"scene_id":"scene1", "character1":"first subject",
"character2":"second subject", "background":"only environment"}]}
Omit character2 and later character fields when those subjects are absent.
\end{tcblisting}

\subsection{Cross-Shot Source Routing}
\label{app:routing-prompts}

\noindent\textbf{Summary.}
Given the extracted slots, a second LLM call selects earlier source slots for
each recall target. Each target character is matched independently and may be
routed to a differently indexed character slot in the source scene; for
example, \texttt{scene2.character1} may refer to
\texttt{scene1.character2}. Empty lists explicitly represent unavailable
references.

\begin{tcblisting}{listing only, breakable, colback=gray!4, colframe=gray!60,
  boxrule=0.4pt, arc=1pt, left=4pt, right=4pt, top=4pt, bottom=4pt,
  listing options={basicstyle=\ttfamily\fontsize{8pt}{10pt}\selectfont, breaklines=true}}
You select source scenes for semantic recall in video generation. For each
recall target scene, assign suitable source scenes from earlier scenes only.
Return JSON only.

# TASK
For each recall target, select earlier source slots for every target character
and for the background. Use the target scene and all eligible earlier scenes
as the basis for the decision.

# INPUTS
## SCENES
scene1: character1={scene1_character1}; character2={scene1_character2};
  background={scene1_background}
scene2: character1={scene2_character1}; background={scene2_background}
...

## RECALL TARGETS
Target scene scene{t}:
  character1={target_character1}
  character2={target_character2}
  ...
  background={target_background}
Only scenes earlier than the target may be used as sources.

# IMMEDIATE PREVIOUS SCENE CONTINUITY RULE
First compare each target with its immediately preceding scene. If character
identity and appearance, character count, background location, and main visual
elements are almost identical, choose that preceding scene for both
the corresponding characterN_sources and background_sources. Do not choose an
older scene unless the preceding scene fails one of these conditions.

# CHARACTER SOURCE REQUIREMENTS
1. Determine the number of distinct people or creatures in the target text.
2. Prefer sources containing only one person or creature. Match identity,
   appearance, clothing, and other static visual features.
3. If several candidates match, select the earliest valid source.
4. Match every target character to the corresponding characterN slot in the
   selected source scene. Source and target slot indices need not be equal.
5. A character source may be empty; never force a non-genuine match.

# BACKGROUND SOURCE REQUIREMENTS
1. Select at most one background source for each target.
2. Match the location and key visual elements, including important objects,
   architecture, environment, lighting, and atmosphere.
3. background_sources may be empty; never force a non-genuine match.

# OUTPUT FORMAT
Return exactly this JSON shape and no additional text:
{"sink_assignments":{"<target_scene_id>":
{"character1_sources":[{"scene_id":"<source_scene_id>",
"source_role":"character2"}],
"background_sources":[{"scene_id":"<source_scene_id>",
"source_role":"background"}]}}}
\end{tcblisting}

\subsection{LLM Inference Cost}
\label{app:llm-cost}
Each LLM request uses approximately 4,000 tokens in total, with an average
reported throughput of about 130 tokens per second (TPS). Dividing this token
budget by the throughput gives a coarse latency estimate of
$4{,}000/130\approx 31$ seconds, or roughly 30 seconds per call.
This estimate is indicative rather than a measured end-to-end latency:
TPS characterizes token generation, whereas the total token budget also
includes input tokens. Actual request latency therefore depends on the
input--output token split and API serving overhead.

\section{Method Details}
\label{app:method-details}

\subsection{Shot Memory Storage}
\label{app:shot-storage}
Each completed shot is registered as one entry in a bounded scene pool. An
entry retains its chronological shot index, original prompt, normalized prompt
feature, and---when semantic routing is enabled---separate character and
background prompt features. The pool is ordered by generation time and discards
its oldest entry when its capacity is exceeded. These lightweight fields locate
historical shots; the visual content is retained separately as K/V memory.

The visual component is stored per transformer layer as a compact pair of key
and value tensors~\citep{attention}. Let $K^{(\ell)}_{s,f,u}$ and $V^{(\ell)}_{s,f,u}$ denote the
key and value at layer $\ell$, candidate latent frame $f$, and spatial token
$u$ of shot $s$. A latent frame contains $U=1560$ spatial tokens in the current
implementation, and each token has 12 attention heads of dimension 128. We use
the \texttt{score\_weighted} mode to aggregate the sampled candidate frames
into one K/V representative per spatial position and layer.
The tensors are detached from the denoising graph and moved to CPU storage;
they are copied back to the active GPU cache only when that shot is selected as
a reference. In LLM-semantic mode, the pipeline additionally stores
cross-attention query summaries, later used to derive slot-specific masks over
the saved K/V tokens. A saved shot therefore comprises retrieval metadata,
compact layer-wise visual K/V memory, and optional semantic-query statistics;
decoded RGB frames are not used as cross-shot memory.

For every layer, the cache accumulates the mean query $\bar Q^{(\ell)}$ and a
bounded candidate K/V buffer while the shot is denoised. The affinity of
candidate frame $f$ at spatial position $u$ is averaged over attention heads,
\begin{equation}
    a^{(\ell)}_{s,f,u}=
    \frac{1}{H}\sum_{h=1}^{H}
    \left\langle \bar Q^{(\ell)}_{h},K^{(\ell)}_{s,f,u,h}\right\rangle,
    \qquad
    \bar Q^{(\ell)}=\frac{1}{N_q}\sum_{i=1}^{N_q}Q^{(\ell)}_i.
    \label{eq:shot-memory-score}
\end{equation}
These scores are normalized over candidate frames,
$\alpha^{(\ell)}_{s,f,u}=\operatorname{softmax}_{f}
(a^{(\ell)}_{s,f,u})$, and used to form weighted K/V summaries~\citep{attention}. To retain a
stable reference, the implementation further blends the weighted summary with
the first sampled candidate frame:
\begin{align}
    \widehat K^{(\ell)}_{s,u}
    &=\lambda K^{(\ell)}_{s,0,u}
      +(1-\lambda)\sum_{f\in\mathcal C_s}
      \alpha^{(\ell)}_{s,f,u}K^{(\ell)}_{s,f,u},\\
    \widehat V^{(\ell)}_{s,u}
    &=\lambda V^{(\ell)}_{s,0,u}
      +(1-\lambda)\sum_{f\in\mathcal C_s}
      \alpha^{(\ell)}_{s,f,u}V^{(\ell)}_{s,f,u},
    \label{eq:shot-memory-score-weighted}
\end{align}
where $\lambda$ is the configured base-frame weight. The same weights are
applied to keys and their co-indexed values, preserving their correspondence in
the compact shot memory
$\mathcal M_s=\{(\widehat K^{(\ell)}_{s,u},
\widehat V^{(\ell)}_{s,u})\}_{\ell=1,u=1}^{L,U}$.

\subsection{Routing Shot Prompts}
\label{app:routing-shot-prompts}
At each shot boundary, the semantic planner uses the LLM-produced~\citep{gemini25pro} slot and
routing results to determine both the historical memory source and the text
used to localize that source. For a target role $r_i$ in shot $t$, the routing
result specifies a source shot $s<t$ and a matched source role
$\pi_t(r_i)$. Importantly, the correspondence is not constrained to preserve
the numerical slot index: a subject represented by
\texttt{character1} in the target may correspond to
\texttt{character2} in the source. The target localization prompt is therefore
read directly from the matched source slot,
\begin{equation}
    P_{\mathrm{tar}}^{(t,i)}
    =P_{s,\pi_t(r_i)},
    \qquad (s,\pi_t(r_i))=\operatorname{LLMRoute}(t,r_i),
    \label{eq:routing-target-prompt}
\end{equation}
rather than being reconstructed from the full target-shot description. This
preserves the identity and appearance wording attached to the actual historical
subject whose memory is being retrieved.

The complementary prompt is constructed from the semantic slots previously
extracted by the LLM from the source shot's own description
(Appendix~\ref{app:slot-prompts}). Let $\mathcal S_s$ denote its non-empty
character and background slots. Non-target slots are the remaining slots
after excluding the matched source slot $\pi_t(r_i)$; their descriptions are
concatenated without an additional LLM call:
\begin{equation}
    P_{\mathrm{oth}}^{(t,i)}
    =\operatorname{Concat}\!\left(
      \{P_{s,r}\}_{r\in\mathcal S_s\setminus\{\pi_t(r_i)\}}
    \right).
    \label{eq:routing-other-prompt}
\end{equation}
For a character target, these slots comprise the background and all other
characters; for a background target, they comprise all characters. Thus,
$P_{\mathrm{tar}}$ describes only the routed component, whereas
$P_{\mathrm{oth}}$ summarizes the remaining non-target content in the same
historical memory.

Both prompts are then encoded to extract attention responses over the same
source memory, using the pretrained query and key projections and their
normalization parameters from the DiT's cross-attention layers. Specifically,
stored source queries are paired with keys projected from each prompt's text
features; scaled query--key scores are averaged over text tokens, attention
heads, and layers to obtain the two spatial response maps used by CSA.
No additional attention parameters are trained.

\subsection{Compressed Window}
\label{app:compressed-window}
This component computes a compact set of latent-token representations from the
completed within-shot history. More precisely, the implementation operates on
the K/V projections of transformer latent tokens~\citep{dit}, rather than averaging or
re-encoding VAE latents~\citep{rombach2022ldm}. Given hidden latent tokens $H^{(\ell)}$, the stored
candidate representation at layer $\ell$ is~\citep{zhang2019rmsnorm}
\begin{equation}
    K^{(\ell)}=\operatorname{RMSNorm}(H^{(\ell)}W_K^{(\ell)}),
    \qquad
    V^{(\ell)}=H^{(\ell)}W_V^{(\ell)}.
    \label{eq:compressed-kv-projection}
\end{equation}
At denoising timestep zero, each newly generated latent frame contributes its
K/V tokens, together with their original temporal and spatial indices, to a
bounded history buffer. The module simultaneously accumulates a calibrated
mean query $\bar Q^{(\ell)}$ over the tokens generated before compression.

The compressed representation is obtained by selecting informative historical
K/V tokens~\citep{zhang2023h2o,li2024snapkv}, not by synthesizing a new latent vector. Each historical token $u$
receives the query-guided score
\begin{equation}
    \rho^{(\ell)}_u=
    \operatorname{Fuse}_{h,\omega}\!\left[
      \operatorname{Re}\!\left(
      \bar Q^{(\ell)}_h\overline{K^{(\ell)}_{u,h}}
      e^{\mathrm{i}\omega\Delta t_u}\right)
      +\operatorname{Amp}\!\left(\bar Q^{(\ell)}_h,K^{(\ell)}_{u,h}\right)
    \right],
    \label{eq:pc-score}
\end{equation}
where $\Delta t_u$ is the token's temporal displacement and
$\operatorname{Fuse}$ aggregates frequency components and attention heads.
The first term measures phase-aware query--key compatibility under temporal
RoPE~\citep{roformer}; the second is an amplitude-compensation term, optionally reduced when
the current query distribution drifts from its calibration. The compressed
representation consists of the highest-scoring eligible history tokens,
\begin{equation}
    \mathcal{I}_{\mathrm{comp}}=
    \operatorname{TopK}_{C}
    \left(\{\rho^{(\ell)}_u\}_{u\in\mathcal{H}^{(\ell)}}\right),
    \qquad
    (K,V)_{\mathrm{comp}}=(K,V)_{\mathcal{I}_{\mathrm{comp}}}.
    \label{eq:pc-selection}
\end{equation}
Here $\mathcal{H}^{(\ell)}$ is the eligible K/V history and $C$ is the fixed
compressed-token budget. The cache limits repeated selection of the same
historical token. It copies the selected K/V pairs together with their original
temporal and spatial indices, so compression preserves the exact values and
positions of selected latent-token representations rather than forming a
blurred temporal average.

\subsection{Semantic Separation}
\label{app:semantic-separation}
Semantic separation determines whether character and background content drawn
from the same historical frame are treated as one shared memory source or as
independent semantic components. Without separation, a source frame selected
for both roles is inserted once and its K/V tokens remain jointly available.
Consequently, subject appearance and scene layout are recalled together, which
is appropriate when the complete historical composition should be preserved
but may also transfer irrelevant background or subject attributes.

With separation enabled, the shared frame is represented by role-specific
views of the same stored K/V tensors. Let $M_{s,r}$ be the semantic mask for
role $r$ in source shot $s$. The routed memory for that role is
\begin{equation}
    \widetilde K_{s,r}=M_{s,r}\odot K_s,
    \qquad
    \widetilde V_{s,r}=M_{s,r}\odot V_s,
    \label{eq:semantic-separated-memory}
\end{equation}
or, equivalently in attention implementation, excluded tokens receive a
strong negative attention bias~\citep{cheng2022mask2former}. Character and background may therefore share
the same source frame while contributing through different spatial token
subsets. The distinction is solely whether co-located semantic content is
kept coupled or separated; it does not change the selected historical frame or
create new visual features.

\subsection{Similarity-Adaptive Decay}
\label{app:similarity-adaptive-decay}
The decay mechanism gradually reduces the influence of retained historical
tokens using a rate bounded by the configured minimum and maximum values. For
each spatial token $u$, the implementation compares the key of the first frame
in the current block, $k^{\mathrm{cur}}_u$, with the corresponding historical
key $k^{\mathrm{hist}}_u$. Their cosine distance is~\citep{manning2008ir}
\begin{equation}
    d_u=1-
    \frac{\langle k^{\mathrm{cur}}_u,k^{\mathrm{hist}}_u\rangle}
    {\lVert k^{\mathrm{cur}}_u\rVert_2
     \lVert k^{\mathrm{hist}}_u\rVert_2},
    \label{eq:decay-cosine-distance}
\end{equation}
where the comparison is averaged over attention heads. The distances are
min--max normalized across the spatial tokens,
\begin{equation}
    \widehat d_u=
    \frac{d_u-d_{\min}}{d_{\max}-d_{\min}},
    \label{eq:decay-normalization}
\end{equation}
with $\widehat d_u=0$ when the distance range is numerically degenerate. Given
the configured bounds $\gamma_{\min}$ and $\gamma_{\max}$, the token-wise decay
rate is obtained by linear interpolation,
\begin{equation}
    \gamma_u=\gamma_{\max}
    -(\gamma_{\max}-\gamma_{\min})\widehat d_u,
    \qquad
    \gamma_u\in[\gamma_{\min},\gamma_{\max}].
    \label{eq:adaptive-decay-rate}
\end{equation}
The historical token weight is then updated multiplicatively as
$w_u\leftarrow\gamma_u w_u$. Hence, tokens that remain similar to the current
content receive a rate near $\gamma_{\max}$ and decay slowly, whereas tokens
with a larger feature discrepancy receive a rate near $\gamma_{\min}$ and are
suppressed more rapidly.

\section{Evaluation Metric Implementation}
\label{app:metrics}

\subsection{Shot-Aware Video Partitioning}
\label{app:metric-preprocessing}

All metrics are evaluated on the native temporal extent of each generated
shot.  In particular, we do not resample, pad, or truncate all methods
to a common number of frames per shot.  Let $V_{n,s}^{(m)}$ be shot $s$ of
video $n$ generated by method $m$, and let $L_{n,s}^{(m)}$ denote its observed
frame count after extraction.  Every frame-based metric operates on the
available index set
\begin{equation}
    \mathcal{F}_{n,s}^{(m)}(\delta)
    = \{20+k\delta \mid 20+k\delta<L_{n,s}^{(m)},\ k\in\mathbb{N}_0\},
    \label{eq:shot-aware-sampling}
\end{equation}
so that its samples reflect the actual duration produced by that method.  The
first 20 frames are skipped to avoid the shot-boundary transient.  When shots
are extracted from a concatenated sequence, the default extractor additionally
removes the first three transition frames from shots 2--6.  These operations
only define shot boundaries; they do not equalize shot lengths.

For cross-shot evaluation, the extractor forms separate subject and background
groups from the recorded routing relations and concatenates the native shot
files in temporal order.  A shot may belong to multiple groups when it reuses
different sources.  Consequently, a grouped video represents one semantic
continuity chain rather than an artificial fixed-length clip.  We use a stride
of 60 frames for within-shot evaluation and 160 frames for grouped cross-shot
evaluation.  A score is reported only when the corresponding sequence contains
the samples required by that metric.

\subsection{Subject and Background Consistency}
\label{app:consistency-metrics}

We use a common region-based procedure for subject and background consistency,
both within shots and across the grouped cross-shot videos described above.
For every sampled RGB frame, YOLO11n~\citep{yolo11ultralytics} detects all COCO person instances
at confidence threshold $0.25$.  Each detection box is supplied to SAM-ViT-B~\citep{kirillov2023sam},
which produces one highest-IoU binary mask per person.  We encode each masked
person crop and the complementary background image using the $[\mathrm{CLS}]$~\citep{dosovitskiy2021vit}
embedding of DINOv2 ViT-B/14~\citep{oquab2023dinov2}, followed by $\ell_2$ normalization.  Person crops
replace non-person pixels inside their bounding rectangles with a neutral value;
for background features, the union of all person masks is filled by the median
colour of the non-person pixels.  Thus, the same encoder and masking protocol
are used for all four settings (within-shot subject/background and cross-shot
subject/background); only the input sequence changes.

Let $\mathbf z_{t,r}$ be the normalized DINOv2 feature of person $r$ in sampled
frame $t$, and let $\mathbf b_t$ be the corresponding background feature.  For
a sequence whose first sampled frame contains reference persons
$\mathcal R$, subject consistency at a later frame $t$ is
\begin{equation}
    c_{\mathrm{sub}}(t)=
    \frac{1}{|\mathcal R|}\sum_{r\in\mathcal R}
    \max_{j\in\mathcal P_t}\mathbf z_{0,r}^{\top}\mathbf z_{t,j},
    \qquad
    C_{\mathrm{sub}}=\frac{1}{|\mathcal T|}
    \sum_{t\in\mathcal T}c_{\mathrm{sub}}(t),
    \label{eq:subject-consistency}
\end{equation}
where $\mathcal P_t$ is the set of detected and segmented persons at $t$ and
$\mathcal T$ contains all later sampled frames.  A later frame without a valid
person mask contributes zero.  If the reference frame has no valid person, the
sequence receives zero; sequences with no later sample are undefined and are
excluded from averaging.  For cross-shot subject groups, the evaluator selects
one reference person per group and prevents duplicate protagonists across groups
of the same video: if its DINOv2 similarity to a previously selected reference
exceeds $0.5$, that group receives zero.

Background consistency is the mean pairwise cosine similarity across all valid
sampled background features,
\begin{equation}
    C_{\mathrm{bg}}=
    \frac{2}{K(K-1)}\sum_{1\leq i<j\leq K}
    \mathbf b_i^{\top}\mathbf b_j,
    \label{eq:background-consistency}
\end{equation}
where $K$ is the number of valid sampled frames.  A sequence with fewer than
two valid background frames is excluded.  Within-shot scores are first averaged
over the available shots of a video; cross-shot scores are first averaged over
the available semantic groups.  Dataset scores are then unweighted means over
valid videos. Runtime failures and undefined sequences are excluded, whereas
the missing-person cases specified above remain penalized by zero.

\subsection{Video Quality and Text Alignment}
\label{app:quality-metrics}

We assess perceptual video quality with two VBench-Long dimensions~\citep{huang2024vbench,huang2025vbenchpp}:
aesthetic quality and imaging quality.  Each method's complete,
unpartitioned output video is passed directly to VBench-Long; no manually
extracted shot is supplied to these two metrics.  VBench-Long applies its
standard long video processing and returns scores for its internally produced
long clips.  For a video with $Q_n$ such clips and dimension $d$, we use
\begin{equation}
    Q_{n,d}=\frac{1}{Q_n}\sum_{q=1}^{Q_n}q_{n,d}^{(q)},
    \qquad
    Q_d=\frac{1}{N}\sum_{n=1}^{N}Q_{n,d}.
    \label{eq:vbenchlong-aggregation}
\end{equation}
The reported imaging-quality value is rescaled from VBench-Long's percentage
output to $[0,1]$ before aggregation.  Both quantities are therefore
whole-video measures, while retaining VBench-Long's native temporal treatment.

For text alignment, the ViCLIP text encoder~\citep{wang2023internvid} has a restricted effective text
length, making concatenation of all six detailed generation prompts unsuitable.
We instead evaluate each extracted shot $V_{n,s}^{(m)}$ independently with the
ordinary VBench ViCLIP \texttt{overall\_consistency} implementation~\citep{huang2024vbench} and its
corresponding short raw prompt $p^{\mathrm{raw}}_{n,s}$ (stored as
\texttt{viclip\_prompt}).  If $a_{n,s}$ denotes the resulting ViCLIP score,
the video- and dataset-level scores are
\begin{equation}
    A_n=\frac{1}{S}\sum_{s=1}^{S}a_{n,s},
    \qquad
    A=\frac{1}{N}\sum_{n=1}^{N}A_n,
    \quad S=6.
    \label{eq:viclip-shot-aggregation}
\end{equation}
This shot-wise design aligns each temporal segment with the instruction that
actually specifies it, while avoiding text truncation.  All reported metrics
are computed from the same JSON-listed videos; result files retain per-video,
per-shot, and per-group scores to permit independent verification.
\begin{figure*}[t]
    \centering
    \includegraphics[page=6,width=\textwidth]{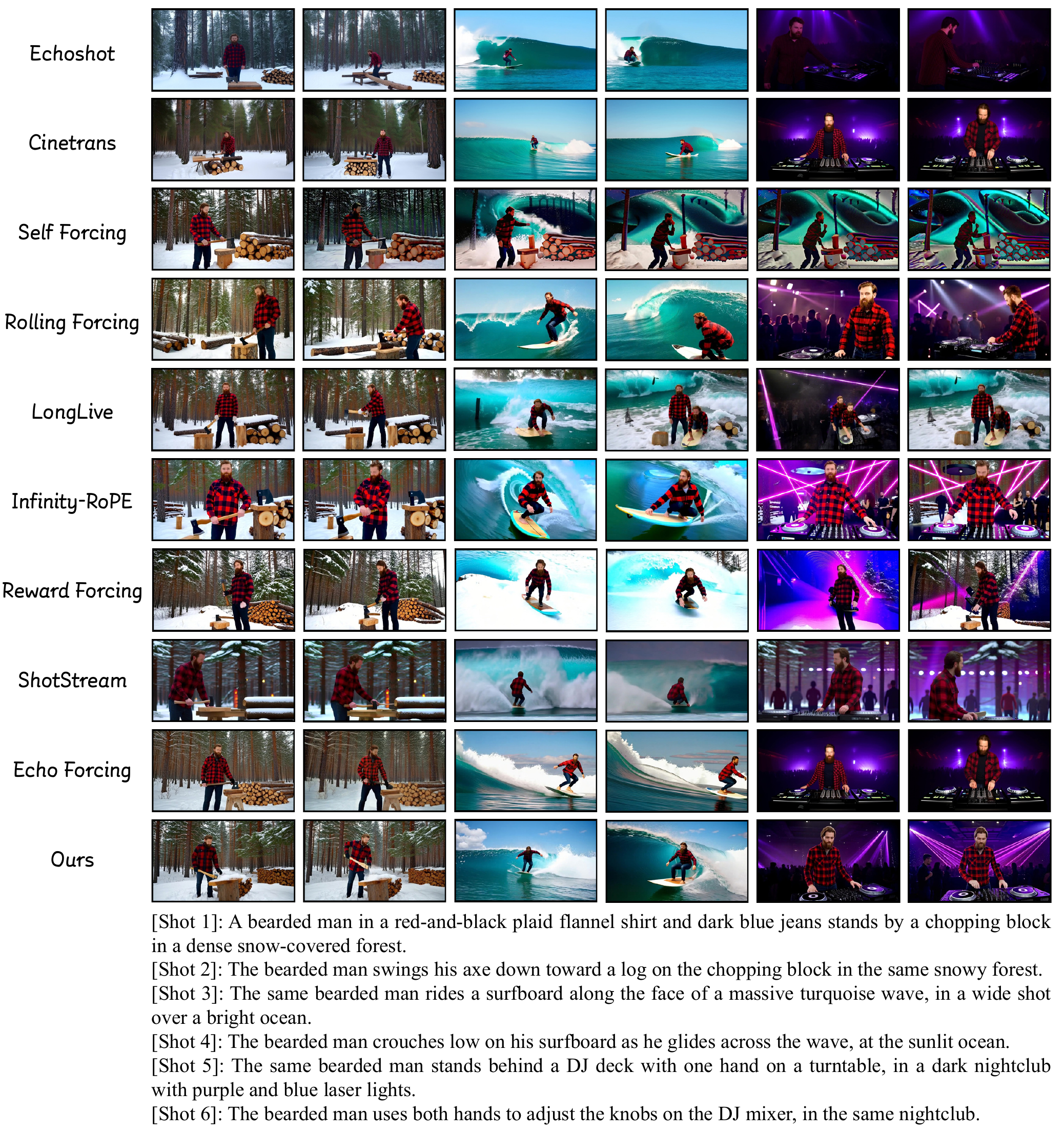}
    \caption{\textbf{Supplementary qualitative comparison in a single-subject
    narrative.} All methods receive the same shot-level
    instructions. The sequence tests preservation of the established subject
    while the setting and event progress change across shots.}
    \label{fig:single-subject-comparison}
\end{figure*}

\section{Additional Qualitative Results}
\label{app:additional-results}

\subsection{Single-Subject Comparison}
\label{app:single-subject-comparison}

The main paper presents a multi-person composition case, where later shots
must jointly recover several identities.  Figure~\ref{fig:single-subject-comparison}
provides the complementary single-subject setting.  It compares all methods
under the same sequence of instructions, isolating whether a method can retain
one established identity while the scene and action evolve.  This setting
separates subject persistence from the additional difficulty of binding
attributes across multiple people.

As shown in Figure~\ref{fig:single-subject-comparison}, Weave Forcing achieves
favorable qualitative results in this controlled setting. Across successive
shots, it preserves the subject's defining appearance while accommodating the
requested changes in scene context and event progression. Compared with the
displayed alternatives, the resulting sequence exhibits fewer conspicuous
identity changes and maintains a more coherent correspondence to the
previously established subject. This observation indicates that the proposed
memory routing is also effective when consistency depends on a single recalled
identity rather than a multi-subject composition.

\label{app:additional-recall-results}

\begin{figure*}[t]
    \centering
    \includegraphics[page=7,width=\textwidth]{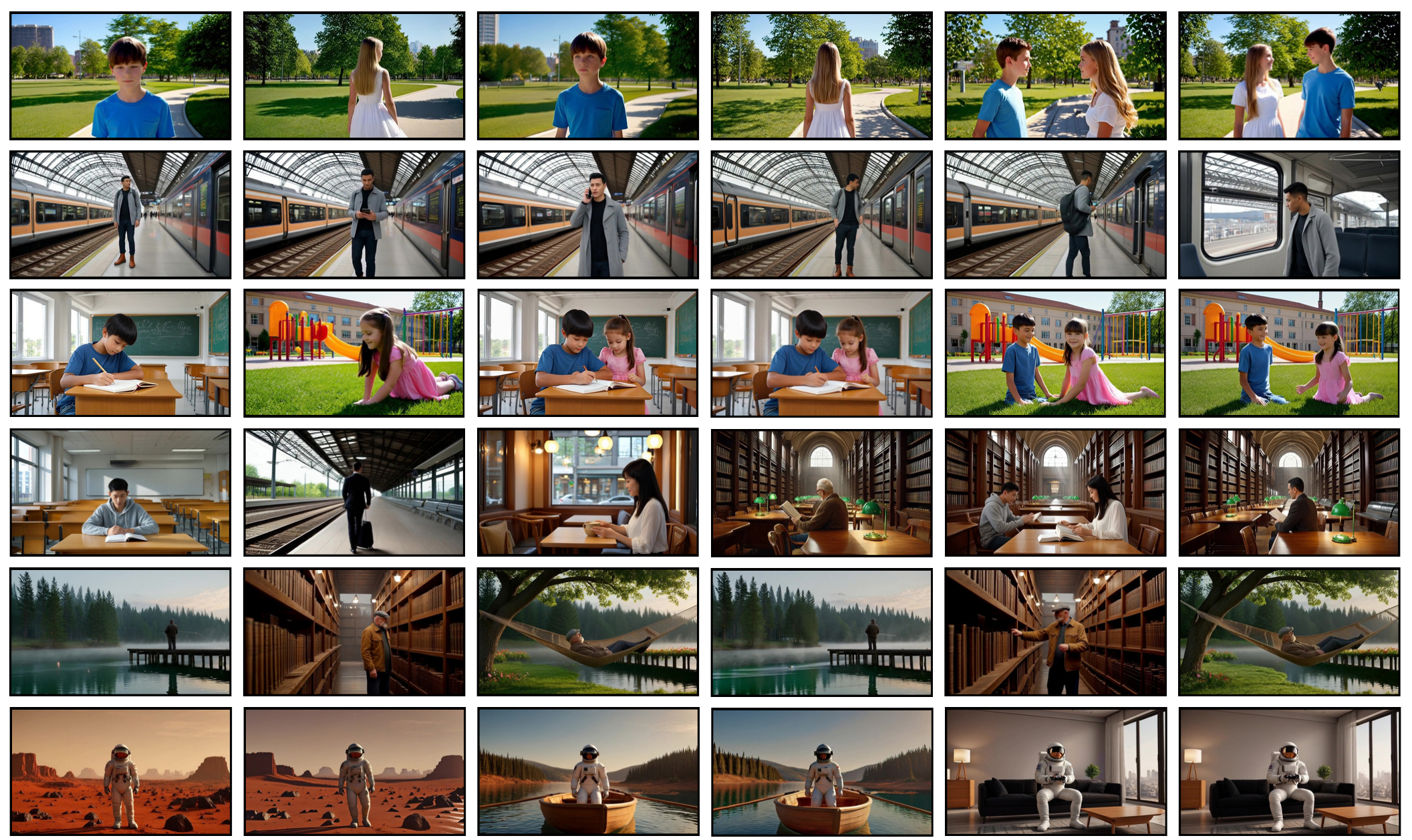}
    \caption{\textbf{Additional Weave Forcing results across diverse interactive
    scenarios.} The examples include scene recall, subject recall, and event
    development, illustrating preservation and controlled evolution of visual
    content across sequential shots.}
    \label{fig:additional-recall-results}
\end{figure*}

Figure~\ref{fig:additional-recall-results} reports additional results for
Weave Forcing without a baseline row, allowing the individual recall patterns
to be inspected at a larger scale.  The examples cover (i) scene recall, in
which a previously established environment is requested again; (ii) subject
recall, in which an earlier character must be preserved under a new context;
and (iii) narrative development, in which the action or event changes while
the relevant visual history remains available.  Together, these cases show the
range of reference-availability and story-progression conditions represented
by the evaluation prompts.

The qualitative results show that Weave Forcing performs consistently across
these heterogeneous scenarios. It can recover previously observed scene
structure, retain subject-specific appearance after a contextual change, and
support continued event development without simply reproducing an earlier
shot. Although the visual requirements differ across the examples, the method
generally preserves the requested historical content while allowing newly
specified elements to evolve, suggesting that its routing mechanism is not
restricted to a single type of cross-shot dependency.

\subsection{Intra-Shot Consistency}
\label{app:intra-shot-results}

\begin{figure*}[t]
    \centering
    \includegraphics[page=8,width=\textwidth]{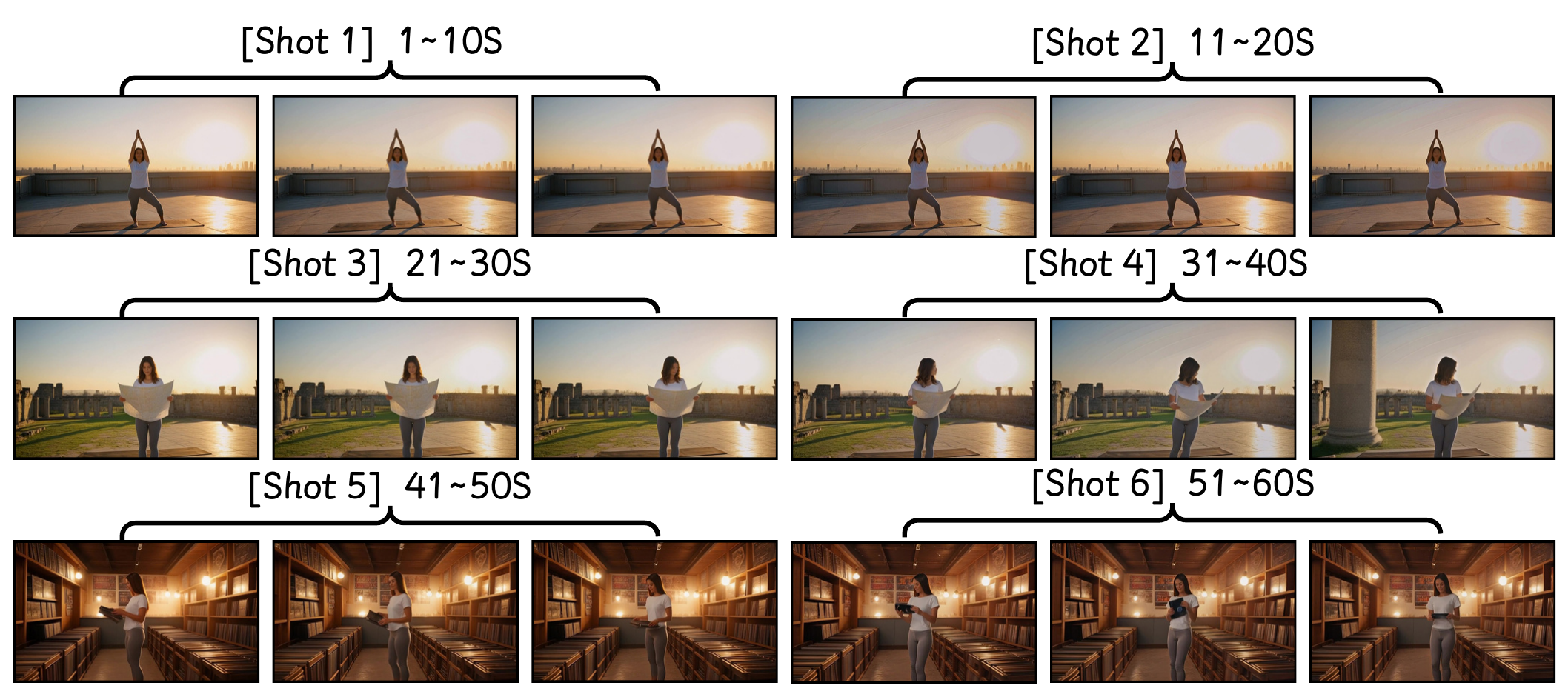}
    \caption{\textbf{Supplementary within-shot consistency results.} Three frames are
    uniformly sampled from each shot to show the persistence of subject
    appearance and local scene content over its temporal extent.}
    \label{fig:intra-shot-results}
\end{figure*}

In addition to cross-shot recall, Figure~\ref{fig:intra-shot-results} shows
the within-shot behavior of the method.  The examples isolate short-range
temporal consistency while a subject and its local environment remain visible
within one shot.  They complement the cross-shot cases above by demonstrating
that the routed historical memory is integrated with, rather than substituted
for, the cache accumulated during the current shot.

For visualization, we uniformly sample three frames from each shot and compare
their subject appearance, local background, and spatial organization. The
sampled frames show satisfactory intra-shot stability, with the principal
visual attributes remaining coherent over the duration of the shot. We
attribute this behavior to two complementary components: the compressed window
retains informative latent K/V tokens from the recent generation history, while
the Self Forcing backbone~\citep{selfforcing} provides a comparatively stable basis for short-range
autoregressive generation. Their combination helps limit local appearance
drift even when routed cross-shot memory is simultaneously introduced.

\section{Limitations and Failure Analysis}
\label{app:limitations}

\begin{figure*}[t]
    \centering
    \includegraphics[page=9,width=\textwidth]{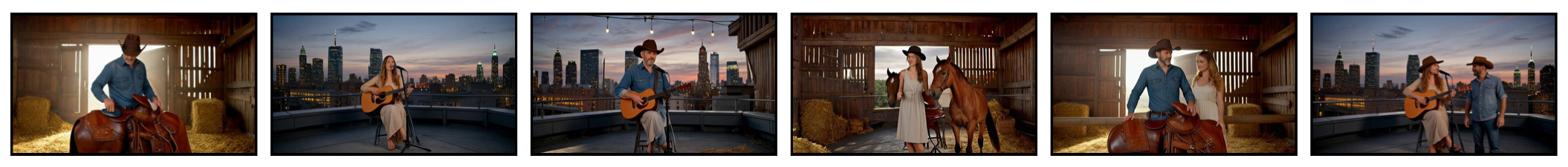}
    \caption{\textbf{Failure case of multi-subject attribute binding.} In the generated
    result, the rear male subject exhibits skirt-like lower clothing, while the
    female subject inherits the male subject's hat. Incomplete semantic masking
    and competition among the many attributes requested by the new prompt can
    cause this cross-subject feature mixing.}
    \label{fig:multi-subject-limitation}
\end{figure*}

Figure~\ref{fig:multi-subject-limitation} illustrates a failure mode in a
crowded multi-subject shot.  Although the generated scene retains both people,
their appearance attributes are not cleanly separated: the lower part of the
rear male subject is rendered with skirt-like clothing, while the female
subject inherits the male subject's hat.  This is an attribute-entanglement~\citep{tang2023daam}
error rather than a complete failure to render either subject.

The failure has two coupled causes.  First, an imperfect semantic mask can
retain spatial tokens from an adjacent subject, leaving insufficient separation
between their respective historical appearances.  Second, the new prompt asks
the generator to attend to many entities and visual attributes simultaneously.
The resulting competition for attention can weaken the slot-specific routing
signal and cause the model to bind an attribute to the wrong person~\citep{chefer2023attendandexcite}.  These
examples suggest that more discriminative instance-level masks and explicit
multi-subject binding are needed when several visually salient subjects must be
recalled and modified in the same shot.

\end{document}